\documentclass[10pt,twocolumn,letterpaper]{article}

\usepackage{cvpr}              

\usepackage{graphicx}
\usepackage{amsmath}
\usepackage{amssymb}
\usepackage{booktabs}

\usepackage{caption}
\usepackage{adjustbox}
\usepackage{wrapfig}
\usepackage{multirow}

\usepackage{algorithm}
\usepackage{algpseudocode}
\usepackage{kotex}
\usepackage[numbers,sort]{natbib}
\usepackage{verbatim}
\usepackage{xcolor}
\usepackage[accsupp]{axessibility}

\newcommand\blfootnote[1]{%
  \begingroup
  \renewcommand\thefootnote{}\footnote{#1}%
  \addtocounter{footnote}{-1}%
  \endgroup
}

\usepackage[pagebackref,breaklinks,colorlinks]{hyperref}

\usepackage[capitalize]{cleveref}
\crefname{section}{Sec.}{Secs.}
\Crefname{section}{Section}{Sections}
\Crefname{table}{Table}{Tables}
\crefname{table}{Tab.}{Tabs.}

\def\cvprPaperID{5279} 
\def\confName{CVPR}
\def\confYear{2022}

\begin{document}

\title{Perception Prioritized Training of Diffusion Models}

\author{Jooyoung Choi$^1$ ~~~~~~~ Jungbeom Lee$^1$ ~~~~~~~ Chaehun Shin$^1$  ~~~~~~~ Sungwon Kim$^1$ \\ Hyunwoo Kim$^3$ ~~~~~~~ Sungroh Yoon$^{1, 2, *}$\\
$^1$ Data Science and AI Laboratory, Seoul National University\\
$^2$ AIIS, ASRI, INMC, ISRC, and Interdisciplinary Program in AI, Seoul National University\\
$^3$ LG AI Research\\
}

\maketitle

\begin{abstract}
\blfootnote{$*$Correspondence to: Sungroh Yoon (sryoon@snu.ac.kr)}
   Diffusion models learn to restore noisy data, which is corrupted with different levels of noise, by optimizing the weighted sum of the corresponding loss terms, i.e., denoising score matching loss. In this paper, we show that restoring data corrupted with certain noise levels offers a proper pretext task for the model to learn rich visual concepts. We propose to prioritize such noise levels over other levels during training, by redesigning the weighting scheme of the objective function. We show that our simple redesign of the weighting scheme significantly improves the performance of diffusion models regardless of the datasets, architectures, and sampling strategies.
\end{abstract}

\section{Introduction}
\label{sec:intro}

Diffusion models~\cite{sohl2015deep,ho2020denoising}, a recent family of generative models, have achieved remarkable image generation performance. Diffusion models have been rapidly studied, as they offer several desirable properties for image synthesis, including stable training, easy model scaling, and good distribution coverage~\cite{nichol2021improved}. Starting from Ho et al.~\cite{ho2020denoising}, recent works~\cite{nichol2021improved,dhariwal2021diffusion,song2020score} have shown that the diffusion models can render high-fidelity images comparable to those generated by generative adversarial networks (GANs)~\cite{goodfellow2014generative}, especially in class-conditional settings, by relying on additional efforts such as classifier guidance~\cite{dhariwal2021diffusion} and cascaded models~\cite{saharia2021image}. However, the unconditional generation of single models still has considerable room for improvement, and performance has not been explored for various high-resolution datasets (e.g., FFHQ~\cite{stylegan}, MetFaces~\cite{karras2020training}) where other families of generative models~\cite{stylegan,vahdat2020nvae,kingma2018glow,esser2021taming,brock2018large} mainly compete.

Starting from tractable noise distribution, a diffusion model generates images by progressively removing noise. To achieve this, a model learns the reverse of the predefined \textit{diffusion process}, which sequentially corrupts the contents of an image with various levels of noise. A model is trained by optimizing the sum of denoising score matching losses~\cite{vincent2011connection} for various noise levels~\cite{song2019generative}, which aims to learn the recovery of clean images from corrupted images. 
Instead of using a simple sum of losses, Ho et al.~\cite{ho2020denoising} observed that their empirically obtained weighted sum of losses was more beneficial to sample quality. Their weighted objective is the current de facto standard objective for training diffusion models~\cite{nichol2021improved,dhariwal2021diffusion,kong2020diffwave,saharia2021image,song2020score}. However, surprisingly, it remains unknown why this performs well or whether it is optimal for sample quality. To the best of our knowledge, the design of a better weighting scheme to achieve better sample quality has not yet been explored.

Given the success of diffusion models with the standard weighted objective, we aim to amplify this benefit by exploring a more appropriate weighting scheme for the objective function. However, designing a weighting scheme is difficult owing to two factors. First, there are thousands of noise levels; therefore, an exhaustive grid search is impossible. Second, it is not clear what information the model learns at each noise level during training, therefore hard to determine the priority of each level.

In this paper, we first investigate what a diffusion model learns at each noise level. Our key intuition is that the diffusion model learns rich visual concepts by solving \textit{pretext tasks} for each level, which is to recover the image from corrupted images. At the noise level where the images are slightly corrupted, images are already available for perceptually rich content and thus, recovering images does not require prior knowledge of image contexts. For example, the model can recover noisy pixels from neighboring clean pixels. Therefore, the model learns imperceptible details, rather than high-level contexts. In contrast, when images are highly corrupted so that the contents are unrecognizable, the model learns perceptually recognizable contents to solve the given pretext task. Our observations motivate us to propose P2 (perception prioritized) weighting, which aims to prioritize solving the pretext task of more important noise levels. We assign higher weights to the loss at levels where the model learns perceptually rich contents while minimal weights to which the model learns imperceptible details.

To validate the effectiveness of the proposed P2 weighting, we first compare diffusion models trained with previous standard weighting scheme and P2 weighting on various datasets. Models trained with our objective are consistently superior to the previous standard objective by large margins. Moreover, we show that diffusion models trained with our objective achieve state-of-the-art performance on CelebA-HQ~\cite{karras2017progressive} and Oxford-flowers~\cite{nilsback2008automated} datasets, and comparable performance on FFHQ~\cite{stylegan} among various types of generative models, including generative adversarial networks (GANs)~\cite{goodfellow2014generative}. We further analyze whether P2 weighting is effective to various model configurations and sampling steps. Our main contributions are as follows:

\begin{itemize}

\item We introduce a simple and effective weighting scheme of training objectives to encourage the model to learn rich visual concepts. 

\item We investigate how the diffusion models learn visual concepts from each noise level.

\item We show consistent improvement of diffusion models across various datasets, model configurations, and sampling steps.
\end{itemize}

\section{Background}
\label{sec:background}

\subsection{Definitions}
\label{sec:background_def}
Diffusion models~\cite{sohl2015deep,ho2020denoising} transform complex data distribution $p_{data}(x)$ into simple noise distribution $\mathcal{N}(0,\mathbf{I})$ and learn to recover data from noise.
The \textit{diffusion process} of diffusion models gradually corrupts data $x_0$ with predefined noise scales $0<\beta _1, \beta _2, ..., \beta_T <1$, indexed by time step $t$.
Corrupted data $x_1,...,x_T$ are sampled from data $x_0\sim p_{data}(x)$, with a diffusion process, which is defined as Gaussian transition:
\begin{equation}\label{eq:forward}
q(x_{t}|x_{t-1})=\mathcal{N}(x_{t};\sqrt{1-\beta _{t}}x_{t-1},\beta _{t}\mathbf{I}).
\end{equation}
Noisy data $x_t$ can be sampled from $x_0$ directly:
\begin{equation}\label{eq:closed}
x_t=\sqrt{\alpha _{t}}x_{0} + \sqrt{1-\alpha _{t}}\epsilon,
\end{equation}
where $\epsilon\sim \mathcal{N}(0,\mathbf{I})$ and $\alpha_{t}:=\prod_{s=1}^t (1-\beta _{s})$. We note that data $x_0$, noisy data $x_1,...,x_T$, and noise $\epsilon$ are of the same dimensionality. 
To ensure $p(x_T)\sim \mathcal{N}(0,\mathbf{I})$ and the reversibility of the diffusion process~\cite{sohl2015deep}, one should set $\beta_t$ to be small and $\alpha_T$ to be near zero. To this end, Ho et al.~\cite{ho2020denoising} and Dhariwal et al.~\cite{dhariwal2021diffusion} use a linear noise schedule where $\beta_t$ increases linearly from $\beta_1$ to $\beta_T$. Nichol et al.~\cite{nichol2021improved} use a cosine schedule where $\alpha_t$ resembles the cosine function.

Diffusion models generate data $x_0$ with the learned \textit{denoising process} $p_\theta(x_{t-1}|x_t)$ which reverses the diffusion process of~\cref{eq:forward}. Starting from noise $x_T\sim \mathcal{N}(0,\mathbf{I})$, we iteratively subtract the noise predicted by noise predictor $\epsilon _\theta$:
\begin{equation}\label{eq:reverse}
x_{t-1}= \frac{1}{\sqrt{1-\beta _t}}(x_t-\frac{\beta_t}{\sqrt{1-\alpha _t}}\epsilon _{\theta}(x_t,t))+\sigma _t z,
\end{equation}
where $\sigma _t^2$ is a variance of the denoising process and $z\sim \mathcal{N}(0,\mathbf{I})$. Ho et al.~\cite{ho2020denoising} used $\beta _t$ as $\sigma _t^2$.

Recent work Kingma et al.~\cite{kingma2021variational} simplified the noise schedules of diffusion models in terms of \textit{signal-to-noise ratio} (SNR). SNR of corrupted data $x_t$ is a ratio of squares of mean and variance from~\cref{eq:closed}, which can be written as:
\begin{equation}\label{eq:snr}
\text{SNR}(t)=\alpha _{t}/(1-\alpha _{t}),
\end{equation}
and thus the variance of noisy data $x_t$ can be written in terms of SNR: $\alpha _t= 1 - 1/(1+\text{SNR}(t))$. We would like to note that SNR($t$) is a monotonically decreasing function.

\begin{figure*}[t!]
  \centering
  \includegraphics[width=1.0\linewidth]{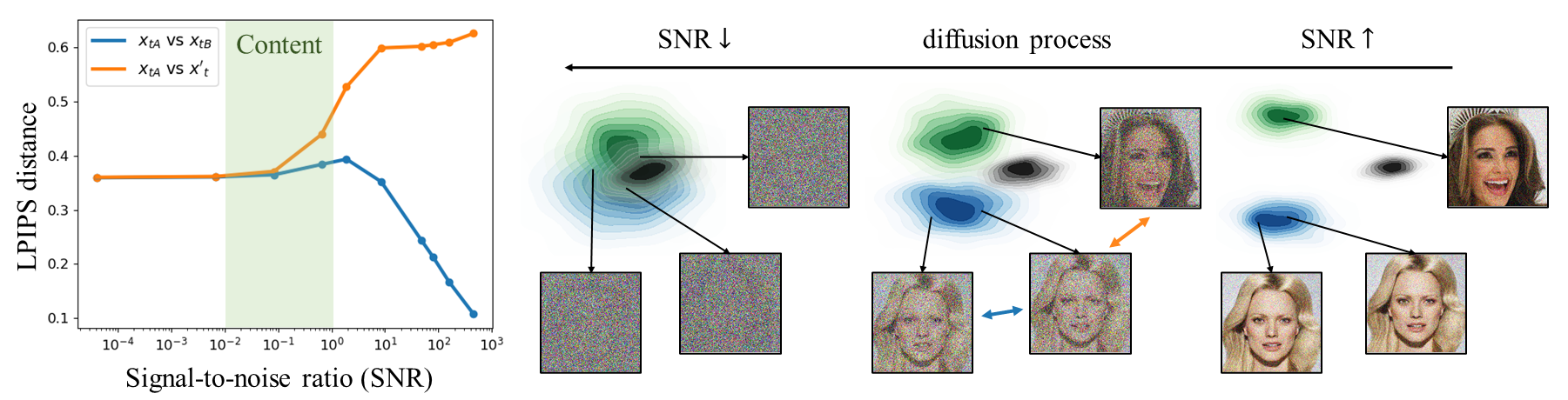}
  \caption{\textbf{Information removal of a diffusion process.} (Left) Perceptual distance of corrupted images as a function of signal-to-noise ratio (SNR). Distances are measured between two noisy images either corrupted from the same image (blue) or different images (orange). 
  We averaged distances measured with 200 random triplets from CelebA-HQ. 
  Perceptually recognizable contents are removed when SNR magnitude is between $10^{-2}$ and $10^0$. 
  (Right) Illustration of the diffusion process.}
  \label{fig:mode}
\end{figure*}

\subsection{Training Objectives}
\label{sec:objective}
The diffusion model is a type of variational auto-encoder (VAE); where the encoder is defined as a fixed \textit{diffusion process} rather than a learnable neural network, and the decoder is defined as a learnable \textit{denoising process} that generates data. Similar to VAE, we can train diffusion models by optimizing a variational lower bound (VLB), which is a sum of denoising score matching losses~\cite{vincent2011connection}: $L_{vlb}=\sum_t L_t$, where weights for each loss term are uniform. For each step $t$, denoising score matching loss $L_t$ is a distance between two Gaussian distributions, which can be rewritten in terms of noise predictor $\epsilon_\theta$ as:
\begin{align}\label{eq:vlb}
L_{t}=~&D_{KL}(q(x_{t-1}|x_t,x_0)~||~p_\theta(x_{t-1}|x_t))  \nonumber \\
=~&\mathbb{E}_{x_0,\epsilon}[\frac{\beta _t}{(1-\beta _t)(1
-\alpha _t)}||\epsilon-\epsilon_\theta(x_t, t)||^2].
\end{align}
Intuitively, we train a neural network $\epsilon _\theta$ to predict the noise $\epsilon$ added in noisy image $x_t$ for given time step $t$.

Ho et al.~\cite{ho2020denoising} empirically observed that the following simplified objective is more beneficial to sample quality:
\begin{equation}\label{eq:simple}
L_{simple}=\sum _t \mathbb{E}_{x_0,\epsilon}[||\epsilon-\epsilon_\theta(x_t, t)||^2].
\end{equation}
In terms of VLB, their objective is $L_{simple}=\sum_t\lambda _t L_t$ with weighting scheme $\lambda _t=(1-\beta _t)(1-\alpha _t)/\beta _t$. In a continuous-time setting, this scheme can be expressed in terms of SNR: 
\begin{equation}\label{eq:simple_snr}
\lambda_t = -1/\text{log-SNR}'(t)= -\text{SNR}(t)/\text{SNR}'(t),
\end{equation}
where $\text{SNR}'(t)=\frac{d\text{SNR}(t)}{dt}$. See appendix for derivations.

While Ho et al.~\cite{ho2020denoising} use fixed values for the variance $\sigma _t$, Nichol et al.~\cite{nichol2021improved} propose to learn it with hybrid objective $L_{hybird}=L_{simple}+c L_{vlb}$, where $c = 1e^{-3}$. They observed that learning $\sigma _t$ enables reducing sampling steps while maintaining the generation performance. 
We inherit their hybrid objective for efficient sampling and modify $L_{simple}$ to improve performance.

\subsection{Evaluation Metrics}
We use FID~\cite{heusel2017gans} and KID~\cite{binkowski2018demystifying} for quantitative evaluations.
FID is well-known to be analogous to human perception~\cite{heusel2017gans} and well-used as a default metric~\cite{karras2020training,dhariwal2021diffusion,stylegan,esser2021taming,parmar2021cleanfid} for measuring generation performances. KID is a well-used metric to measure performance on small datasets~\cite{karras2020training,karras2021alias,parmar2021cleanfid}. However, since both metrics are sensitive to the preprocessing~\cite{parmar2021cleanfid}, we use a correctly implemented library~\cite{parmar2021cleanfid}. We compute FID and KID between the generated samples and the entire training set. We measured final scores with 50k samples and conducted ablation studies with 10k samples for efficiency, following~\cite{dhariwal2021diffusion}. We denote them as FID-50k and FID-10k respectively.

\section{Method}
\label{sec:method}
We first investigate what the model learns at each diffusion step in~\cref{sec:method_3.1}. Then, we propose our weighting scheme in~\cref{sec:method_3.2}. We provide discussions on how our weighting scheme is effective in~\cref{sec:method_3.3}.

\begin{figure*}[t!]
  \centering
  \includegraphics[width=1.0\linewidth]{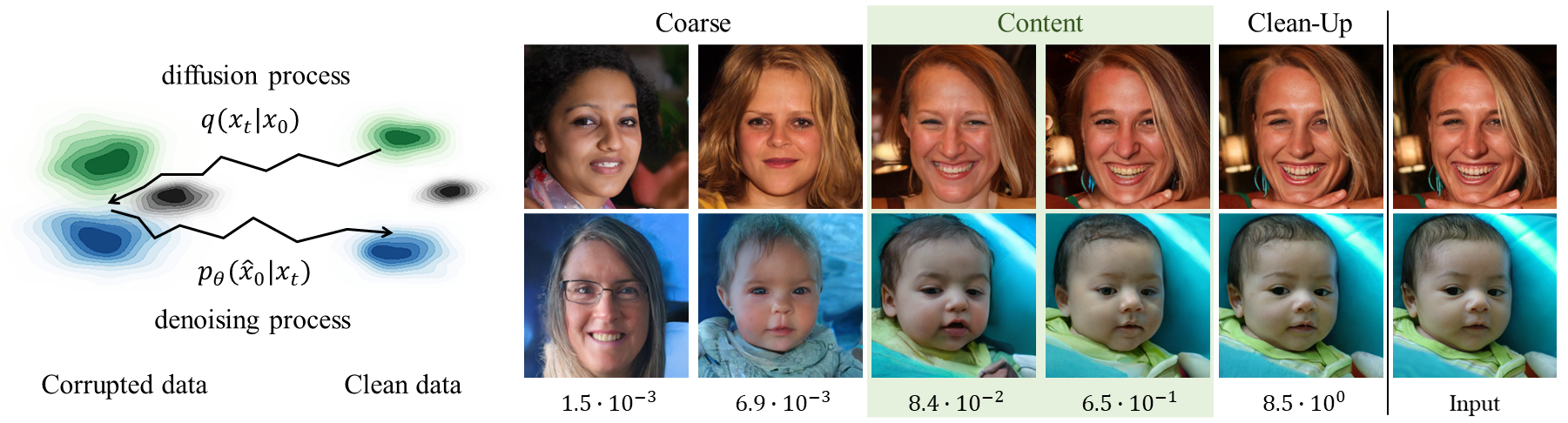}
  \caption{\textbf{Stochastic reconstruction.} (Left) Illustration of reconstruction, where sample are obtained from full sampling chain. 
  (Right) Reconstructions $\hat{x}_0$ with input images $x_0$ on the rightmost column and SNR of $x_t$ on the bottom. Samples in the 1st, 2nd columns share only the coarse attributes (e.g., global color structure) with the input. The 3rd, 4th columns share perceptually discriminative contents with the input. 5th column are almost identical to the input, suggesting that the model learns imperceptible details when SNRs are large.}
  \label{fig:recon}
\end{figure*}

\subsection{Learning to Recover Signals from Noisy Data}
\label{sec:method_3.1}
Diffusion models learn visual concepts by solving \textit{pretext task} at each noise level, which is to recover signals from corrupted signals.
More specifically, the model predicts the noise component $\epsilon$ of a noisy image $x_t$, where the time step $t$ is an index of the noise level. While the output of diffusion models is noise, other generative models (VAE, GAN) directly output images. Because noise does not contain any content or signals, it is difficult to understand how the noise predictions contribute to learning rich visual concepts. Such nature of diffusion models arises the following question: \textit{what information does the model learn at each step during training?} 

\textbf{Investigating diffusion process.  }
We first investigate the predefined diffusion process to explore what the model can learn from each noise level. 
Let say we have two different clean images $x_0$, $x'_0$ and three noisy images $x_{tA},~x_{tB}\sim q(x_t|x_0)$, $x'_t\sim q(x_t|x'_0)$, where $q$ is the diffusion process. In~\cref{fig:mode} (left), we measure perceptual distances (LPIPS~\cite{zhang2018unreasonable}) in two cases: the distance between $x_{tA}$ and $x_{tB}$ (blue line), which share the same $x_0$, and the distance between $x_{tA}$ and $x'_t$ (orange line), which were synthesized from different images $x_0$ and $x'_0$. We present the distances of the two cases as functions of the \textit{signal-to-noise ratio} (SNR) introduced in~\cref{eq:snr}, which characterizes the noise level at each step. To briefly review, SNR decreases through the diffusion process, as shown in~\cref{fig:mode} (right), and increases through the denoising process.

The early steps of the diffusion process have large SNRs, which indicates invisibly small noise; thus, noisy images $x_t$ retain a large amount of contents from the clean image $x_0$. Therefore, in the early steps, $x_{tA}$ and $x_{tB}$ are perceptually similar, while $x_{tA}$ and $x'_t$ are perceptually different, as shown by the large SNR side in~\cref{fig:mode} (left). A model can recover signals without understanding holistic contexts, as perceptually rich signals are already prepared in the image. Thus the model will learn only imperceptible details by solving recovery tasks when SNR is large. 

In contrast, the late steps have small SNRs, indicating a sufficiently large noise to remove the contents of $x_0$. Therefore, distances of both cases start to converge to a constant value, as the noisy images become difficult to recognize the high-level contents. It is shown in the small SNR side in~\cref{fig:mode} (left). Here, a model needs prior knowledge to recover signals because the noisy images lack recognizable content. We argue that the model will learn perceptually rich contents by solving recovery tasks when SNR is small.

\textbf{Investigating a trained model.  }
We would like to verify the aforementioned discussions with a trained model.
Given an input image $x_0$, we first perturb it to $x_t$ using a diffusion process $q(x_t|x_0)$ and reconstruct it with the learned denoising process $p_\theta(\hat{x}_0|x_t)$, as illustrated in~\cref{fig:recon} (left). When $t$ is small, the reconstruction $\hat{x}_0$ will be highly similar to the input $x_0$ as the diffusion process removes a small amount of signals, while $\hat{x}_0$ will share less content with $x_0$ when $t$ is large. In \cref{fig:recon} (right), we compare $x_0$ and $\hat{x}_0$ among various $t$ to show how each step contributes to the sample. 
Samples in the first two columns share only coarse features (e.g., global color scheme) with the input on the rightmost column, whereas samples in the third and fourth columns share perceptually discriminative contents. This suggests that the model learns coarse features when the SNR of step $t$ is smaller than $10^{-2}$ and the model learns the content when SNR is between $10^{-2}$ and $10^0$. When the SNR is larger than $10^0$ (fifth column), reconstructions are perceptually identical to the inputs, suggesting that the model learns imperceptible details that do not contribute to perceptually recognizable contents.

Based on the above observations, we hypothesize that diffusion models learn coarse features (e.g., global color structure) at steps of small SNRs ($0\text{--}10^{-2}$), perceptually rich contents at medium SNRs ($10^{-2}\text {--}10^0$), and remove remaining noise at large SNRs ($10^{0}\text{--}10^4$). 
According to our hypothesis, we group noise levels into three stages, which we term \textit{coarse}, \textit{content}, and \textit{clean-up} stages.

\subsection{Perception Prioritized Weighting}
\label{sec:method_3.2}

In the previous section, we explored what the diffusion model learns from each step in terms of SNR. We discussed that the model learns coarse features (e.g., global color structure), perceptually rich contents, and to clean up the remaining noise at three groups of noise levels. We pointed out that the model learns imperceptible details at the clean-up stage. In this section, we introduce Perception Prioritized (P2) weighting, a new weighting scheme for the training objective, which aims to prioritize learning from more important noise levels.

We opt to assign minimal weights to the unnecessary clean-up stage thereby assigning relatively higher weights to the rest. In particular, we aim to emphasize training on the content stage to encourage the model to learn perceptually rich contexts. 
To this end, we construct the following weighting scheme:
\begin{equation}\label{eq:ours}
\lambda'_t = \frac{\lambda_t}{(k+\text{SNR}(t))^{\gamma}},
\end{equation}
where $\lambda_t$ is the previous standard weighting scheme (\cref{eq:simple_snr}) and $\gamma$ is a hyperparameter that controls the strength of down-weighting focus on learning imperceptible details. $k$ is a hyperparameter that prevents exploding weights for extremely small SNRs and determines sharpness of the weighting scheme. While multiple designs are possible, we show that even the simplest choice (P2) outperforms the standard scheme $\lambda_t$. Our method is applicable to existing diffusion models by replacing $\sum_t\lambda _t L_t$ with $\sum_t\lambda' _t L_t$.

In fact, our weighting scheme $\lambda'_t$ is a generalization of the popularly used~\cite{nichol2021improved,dhariwal2021diffusion,kong2020diffwave,song2020score} weighting scheme $\lambda_t$ of Ho et al.~\cite{ho2020denoising} (\cref{eq:simple_snr}), where $\lambda'_t$ arrives at $\lambda_t$ when $\gamma=0$. We refer to $\lambda_t$ as the \textit{baseline} herein. 

\begin{figure*}[t!]
  \centering
  \includegraphics[width=1.0\linewidth]{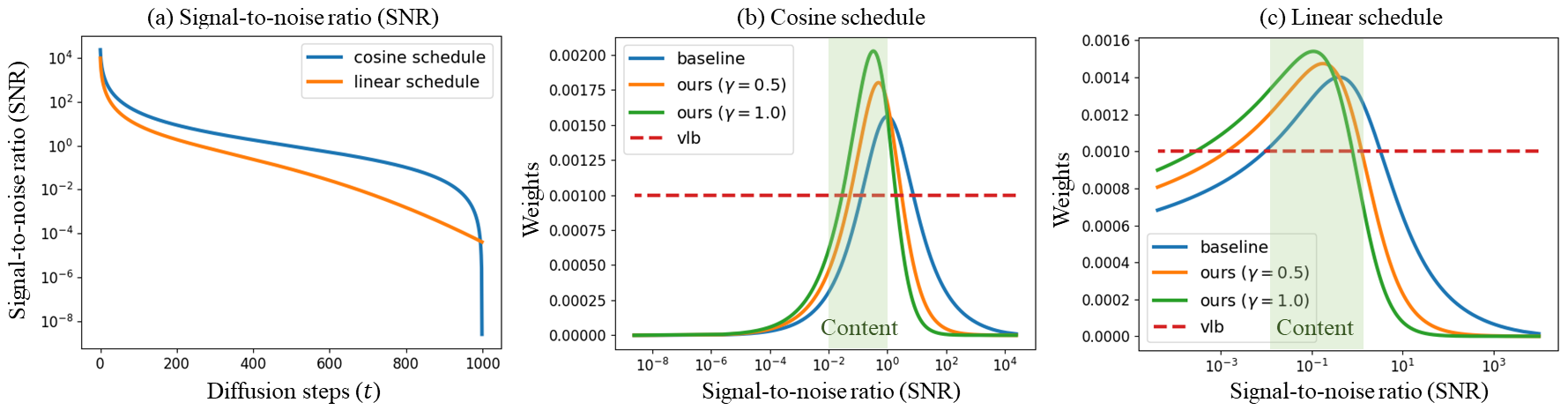}
  \caption{\textbf{Weighting schemes.} (Left) Signal-to-noise ratio (SNR) of linear and cosine noise schedules for reference. (Middle) Weights of our P2 weighting and the baseline with a cosine schedule. (Right) Weights of P2 weighting and the baseline with a linear schedule. Compared to the baseline, P2 weighting suppresses weights for large SNRs, where the model learns imperceptible details.}
  \label{fig:snr_weight}
\end{figure*}

\begin{figure}[t!]
  \centering
  \includegraphics[width=0.9\linewidth]{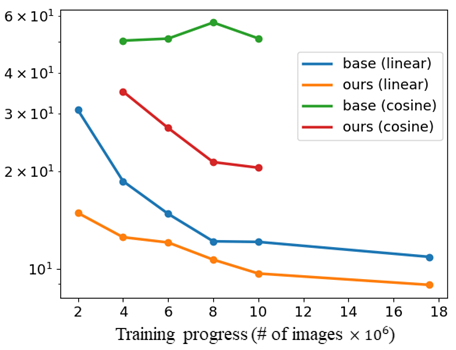}
  \caption{\textbf{Comparison of FID-10k through training on FFHQ.} P2 weighting consistently improves performance for both linear and cosine schedules. Training progress refers to the number of images seen by the model. Samples are generated with 250 steps.}
  \label{fig:ablation}
\end{figure}

\begin{figure*}[t!]
  \centering
  \includegraphics[width=1.0\linewidth]{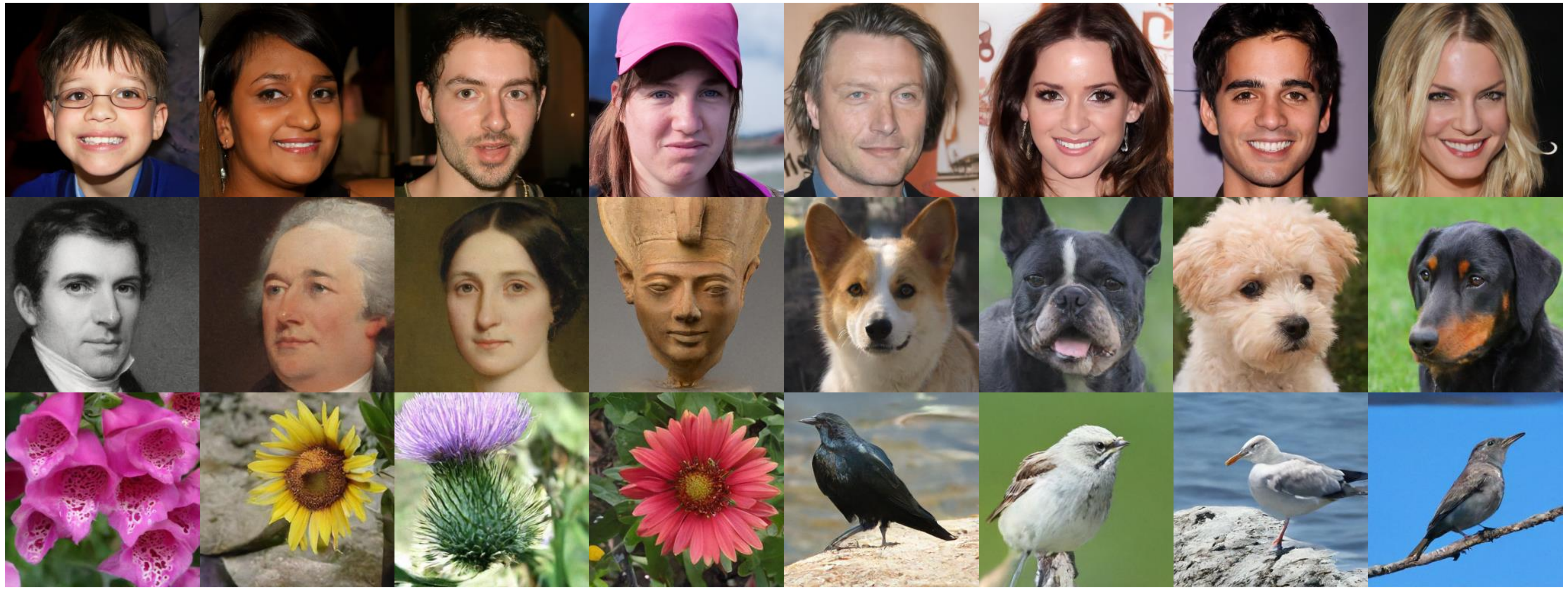}
  \caption{Samples generated by our models trained on several datasets (FFHQ, CelebA-HQ, MetFaces, AFHQ-Dogs, Oxford Flowers, CUB Bird) at 256$\times$256 resolution. See appendix for more samples.}
  \label{fig:teaser}
\end{figure*}

\subsection{Effectiveness of P2 Weighting}
\label{sec:method_3.3}
Prior works~\cite{ho2020denoising,nichol2021improved} empirically suggest that the baseline objective $\sum_t\lambda _t L_t$ offers a better inductive bias for sample quality than the VLB objective $\sum_t L_t$, which does not impose any inductive bias during training. \cref{fig:snr_weight} exhibits $\lambda'_t$ and $\lambda_t$ for both linear~\cite{ho2020denoising} and cosine~\cite{nichol2021improved} noise schedules, which are explained in~\cref{sec:background_def}, indicating that both weighting schemes focus training on the content stage the most and the cleaning stage the least. The success of the baseline weighting is in line with our previous hypothesis that models learn perceptually rich content by solving pretext tasks at the content stage.

However, despite the success of the baseline objective, we argue that the baseline objective still imposes an undeserved focus on learning imperceptible details and prevents from learning perceptually rich content. \cref{fig:snr_weight} shows that our $\lambda'_t$ further suppresses the weights for the cleaning stage, which relatively uplifts the weights for the coarse and the content stages. To visualize relative changes of weights, we exhibit normalized weighting schemes.
\cref{fig:ablation} supports our method in that FID of the diffusion model trained with our weighting scheme ($\gamma = 1$) beats the baseline for both linear and cosine schedules throughout the training. 

Another notable result from \cref{fig:ablation} is that the cosine schedule is inferior to the linear schedule by a large margin, although our weighting scheme improves the FID by a large gap. 
\cref{eq:vlb} indicates that the weighting scheme is closely related to the noise schedule.
As shown in~\cref{fig:snr_weight}, the cosine schedule assigns smaller weights to the content stage compared with the linear schedule. We would like to note that designing weighting schemes and noise schedules are correlated but not equivalent, as the noise schedules affects both weights and MSE terms.


To summarize, our P2 weighting provides a good inductive bias for learning rich visual concepts, by uplifting weights at the coarse and the content stages, and suppressing weights at the clean-up stage.

\subsection{Implementation}
We set $k$ as 1 for easy deployment, because $1/(1+\text{SNR}(t))=1-\alpha_t$, as discussed in~\cref{sec:background_def}. We set $\gamma$ as either 0.5 and 1. We empirically observed that $\gamma$ over 2 suffers noise artifacts in the sample because it assigns almost zero weight to the clean-up stage. We set $T=1000$ for all experiments. We implemented the proposed approach on top of ADM~\cite{dhariwal2021diffusion}, which offers well-designed architecture and efficient sampling. We use lighter version of ADM through our experiments. Our code and models are available\footnote{\url{https://github.com/jychoi118/P2-weighting}}.

\section{Experiment}
\label{sec:experiment}
We start by exhibiting the effectiveness of our new training objective over the baseline objective in~\cref{sec:experiment_baseline}. Then, we compare with prior literature of various types of generative models in~\cref{sec:experiment_literature}. Finally, we conduct analysis studies to further support our method in~\cref{sec:experiment_analysis}. Samples generated with our models are shown in~\cref{fig:teaser}.

\begin{table}[]
\centering
\begin{tabular}{l|r|cc|cc}
\hline
                        & \multicolumn{1}{l|}{}     & \multicolumn{2}{c|}{FID-50k$\downarrow$}                             & \multicolumn{2}{c}{KID-50k$\downarrow$}                             \\ \hline
Dataset                 & \multicolumn{1}{l|}{Step} & \multicolumn{1}{c}{Base} & \multicolumn{1}{c|}{Ours} & \multicolumn{1}{c}{Base} & \multicolumn{1}{c}{Ours} \\ \hline
\multirow{2}{*}{FFHQ}   & 1000                      &             7.86         &         \textbf{6.92}      &       3.85          &      \textbf{3.46}        \\
                        & 500                       &            8.41          &         \textbf{6.97}     &       4.48        &         \textbf{3.56}         \\  \hline
\multirow{2}{*}{CUB}    & 1000                      &            9.60   &         \textbf{6.95}         &      3.49              &         \textbf{2.38}         \\
                        & 250                       &              10.26     &         \textbf{6.32}  &        4.06        &     \textbf{1.93}         \\  \hline
\multirow{2}{*}{AFHQ-D} & 1000                      &       12.47       &         \textbf{11.55}     &         4.79     &  \textbf{4.10}         \\
                        & 250                       &         12.95      &         \textbf{11.66}         &         5.25         &         \textbf{4.20}         \\  \hline
Flowers                 & 250                       &             20.01     &         \textbf{17.29}   &       16.8       &     \textbf{14.8}         \\
MetFaces                & 250                       &              44.34       &         \textbf{36.80}      &          22.1       &      \textbf{17.6}         \\ \hline
\end{tabular}
\caption{\textbf{Quantitative comparison.} Diffusion models trained with our weighting scheme achieve consistent improvement over the baseline at various datasets and sampling steps, in terms of both FID and KID ($\times 10^3$).}
\label{table:fid_various}
\end{table}


\subsection{Comparison to the Baseline}
\label{sec:experiment_baseline}
\textbf{Quantitative comparison.  } 
We trained diffusion models by optimizing training objectives with both baseline and our weighting scheme on FFHQ~\cite{stylegan}, AFHQ-dog~\cite{choi2020stargan}, MetFaces~\cite{karras2020training}, and CUB~\cite{wah2011caltech} datasets. These datasets contain approximately 70k, 50k, 1k, and 12k images respectively.  We resized and center-cropped data to 256$\times$256 pixels, following the pre-processing performed by ADM~\cite{dhariwal2021diffusion}.

\cref{table:fid_various} shows the results. Our method consistently exhibits superior performance to the baseline in terms of FID and KID. The results suggest that our weighting scheme imposes a good inductive bias for training diffusion models, regardless of the dataset. Our method outperforms the baseline by a large margin especially on MetFaces, which contains only 1k images. Hence, we assume that wasting the model capacity on learning imperceptible details is very harmful when training with limited data.

\begin{figure*}[t!]
  \centering
  \includegraphics[width=1.0\linewidth]{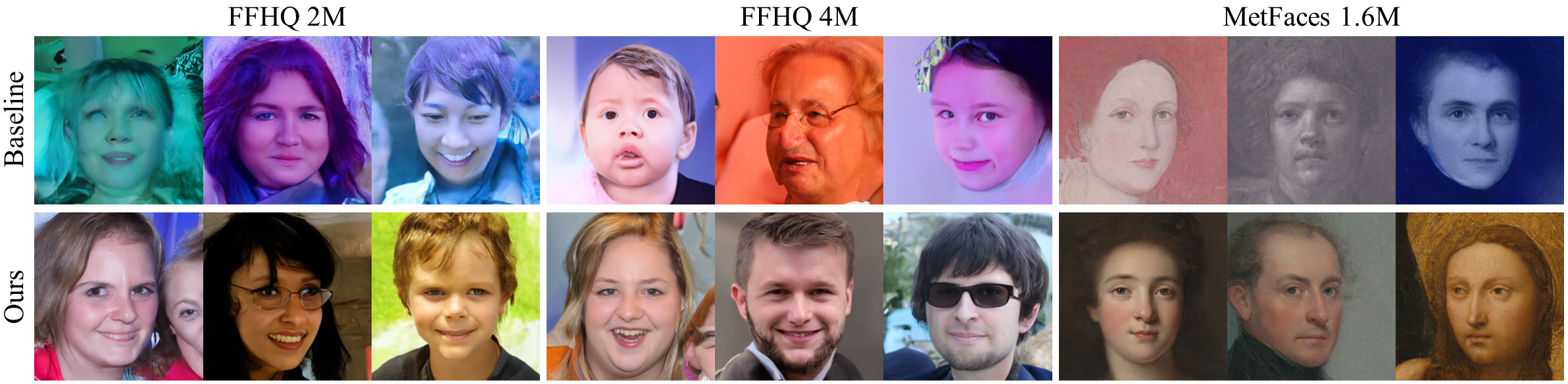}
  \caption{\textbf{Qualitative comparison.} Uncurated samples generated during training. The number of images seen for training displayed on the top. We observed that the baseline suffer color shift problem at early iterations of the training (FFHQ) or even at the convergence (MetFaces). The baseline weighting scheme fails to focus on global coherence and waste model capacity on the imperceptible details.}
  \label{fig:quality_compare}
\end{figure*}

\textbf{Qualitative comparison.  }
We observe that diffusion models trained with the baseline objective are likely to suffer color shift artifacts, as shown in~\cref{fig:quality_compare}. We assume that the baseline training objective unnecessarily focuses on the imperceptible details; therefore, it fails to learn global color schemes properly. In contrast, our objective encourages the model to learn global and holistic concepts in a dataset.

\subsection{Comparison to the Prior Literature}
\label{sec:experiment_literature}
We compare diffusion models trained with our method to existing models on FFHQ~\cite{stylegan}, Oxford flowers~\cite{nilsback2008automated}, and CelebA-HQ~\cite{karras2017progressive} datasets, as shown in~\cref{table:fid_ffhq2}. We use 256$\times$256 resolutions for all datasets. We achieve state-of-the-art FIDs on the Oxford Flowers and CelebA-HQ datasets. While our models are trained with $T=1000$, we already achieve state-of-the-art with reduced sampling steps; 250 and 500 steps respectively. On FFHQ, we achieve a superior result to most models except StyleGAN2~\cite{karras2020analyzing}, whose architecture was carefully designed for FFHQ. We note that our method brought the diffusion model closer to the state-of-the-art, and scaling model architectures and sampling steps will further improve the performance.

\begin{table}[]
\centering
\begin{tabular}{lllr}
\hline
Dataset                                                                   & Method   & Type & FID$\downarrow$ \\ \hline
\multirow{10}{*}{FFHQ}                                                     & $\text{BigGAN}_{\text{~ICLR'19}}$~\cite{brock2018large}  & GAN  &   12.4  \\
                                                                          & $\text{UNet GAN}_{\text{~CVPR'20}}$~\cite{schonfeld2020u}  &  GAN  &  10.9   \\
                                                                          & $\text{StyleGAN2}_{\text{~CVPR'20}}$~\cite{karras2020analyzing}   & GAN  & \textbf{3.73}  \\
                                                                          & $\text{NVAE}_{\text{~NeurIPS'20}}$~\cite{vahdat2020nvae} &  VAE   &   26.02  \\
                                                                          & $\text{VDVAE}_{\text{~ICLR'21}}$~\cite{child2020very}  &  VAE  &   33.5  \\
                                                                          & $\text{VQGAN}_{\text{~CVPR'21}}$\cite{esser2021taming}  &  GAN+AR &   9.6  \\
                                                                          & $\text{D2C}_{\text{~NeurIPS'21}}$~\cite{sinha2021d2c} &  Diff  &   13.04  \\
                                                                          & Baseline (500 step)   &  Diff  &  8.41   \\
                                                                          & P2 (500 step)     &  Diff &  6.97   \\
                                                                          & P2 (1000 step)     &  Diff &  \underline{6.92}  \\ \hline
\multirow{5}{*}{\begin{tabular}[c]{@{}l@{}}Oxford\\ Flower\end{tabular}} & $\text{PGGAN}_{\text{~ICLR'18}}$~\cite{karras2017progressive} &  GAN  &  64.40   \\
                                                                          & $\text{StyleGAN1}_{\text{~CVPR'19}}$~\cite{stylegan}      &  GAN   &  64.70   \\
                                                                          & $\text{MSG-GAN}_{\text{~CVPR'20}}$~\cite{karnewar2020msg}     & GAN   & \underline{19.60}  \\
                                                                          & Baseline (250step)      &   Diff  &  20.01   \\ 
                                                                          & P2 (250step)      &   Diff  &  \textbf{17.29}   \\ \hline
\multirow{8}{*}{\begin{tabular}[c]{@{}l@{}}CelebA\\ -HQ\end{tabular}}     & $\text{PGGAN}_{\text{~ICLR'18}}$~\cite{karras2017progressive}   & GAN   &  8.03   \\
                                                                          & $\text{GLOW}_{\text{~NeurIPS'18}}$~\cite{kingma2018glow}        &   Flow &   68.93  \\
                                                                          & $\text{ALAE}_{\text{~CVPR'20}}$~\cite{pidhorskyi2020adversarial}   &  GAN  &   19.21  \\
                                                                          & $\text{NVAE}_{\text{~NeurIPS'20}}$~\cite{vahdat2020nvae}   &  VAE &   29.76  \\
                                                                          & $\text{VAEBM}_{\text{~ICLR'21}}$~\cite{xiao2020vaebm}   &  VAE+EM  &  20.38   \\
                                                                          & $\text{VQGAN}_{\text{~CVPR'21}}$~\cite{esser2021taming}  &    GAN+AR  &   10.70  \\
                                                                          & $\text{LSGM}_{\text{~NeurIPS'21}}$~\cite{vahdat2021score}    &  VAE+Diff &  \underline{7.22} \\
                                                                          & P2 (500step)   &   Diff   &  \textbf{6.91} \\ \hline
\end{tabular}
\caption{\textbf{Comparison to prior literature.} FFHQ results reproduced from~\cite{parmar2021cleanfid,esser2021taming,sinha2021d2c}, Oxford Flower from~\cite{karnewar2020msg}, and CelebA-HQ from from~\cite{vahdat2021score}. Except for GANs, we achieve superior results.}
\vspace{-1em}
\label{table:fid_ffhq2}
\end{table}

\subsection{Analysis}
\label{sec:experiment_analysis}
In this section, we analyze whether our weighting scheme is robust to the model configurations, number of sampling steps, and sampling schedules.

\textbf{Model configuration matters?  }
Previous experiments are conducted using our default model for fair comparisons. Here, we show that P2 weighting is effective regardless of the model configurations. \cref{table:architecture} shows that our method achieves consistently superior performance to the baseline for various configurations. We investigated for following variations: replacing the BigGAN~\cite{brock2018large} residual block with the residual block of Ho et al.~\cite{ho2020denoising}, removing self-attention at 16$\times$16, using two BigGAN~\cite{brock2018large} residual blocks, and training our default model with a learning rate of $2.5\times10^{-5}$. Our default model contains a single BigGAN~\cite{brock2018large} residual block and is trained with a learning rate $2\times10^{-5}$. Our weighting scheme consistently improves FID and KID by a large margin, across various model configurations. Our method is especially effective when the self-attention is removed ((c)), indicating that P2 encourages learning global dependency.
\vspace{0.5em}

\textbf{Sampling step matters?  }
We trained our models on 1000 diffusion steps following the convention of previous studies. However, it requires more than 10 min to generate a high-resolution image with a modern GPU. Nichol et al.~\cite{nichol2021improved} have shown that their sampling strategy maintains performance even when reducing the sampling steps. They also observed that using the DDIM~\cite{song2020denoising} sampler was effective when using 50 or fewer sampling steps.

\begin{figure}[t!]
  \centering
  \includegraphics[width=0.9\linewidth]{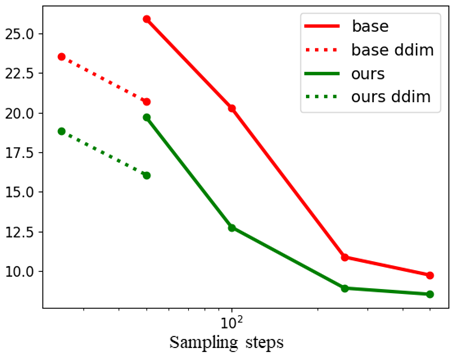}
  \caption{\textbf{Reducing sampling steps.} FID as a function of sampling steps. Our method is superior to the baseline regardless of the sampling steps. Samples generated following~\cite{nichol2021improved} and DDIM~\cite{song2020denoising}. Models trained on FFHQ dataset.}
  \label{fig:speed}
\end{figure}

\cref{fig:speed} shows the FID scores of various sampling steps with models trained on the FFHQ. A model trained with our weighting scheme consistently outperforms the baseline by considerable margins. It should be noted that our weighting scheme consistently achieves better performance with half the number of sampling steps required by the baseline.

\textbf{Why not schedule sampling steps?  } 
In addition to the consistent improvement across various sampling steps, we sweep over the sampling steps in~\cref{table:sweep}. Sweeping sampling schedules slightly improves FID and KID but does not reach our improvement. Our method is more effective compared to scheduling sampling steps, as we improve the \textit{model training}, which benefits predictions at all steps.

\begin{table}[]
\centering
\begin{tabular}{l|cc|cc}
\hline
                    & \multicolumn{2}{c|}{FID-10k$\downarrow$} & \multicolumn{2}{c}{KID-10k$\downarrow$} \\ \hline
       & Base     & Ours     & Base     & Ours     \\ \hline
(a)  &    46.80    &    $\text{41.93}_{~\textcolor{cyan}{(-4.87)}}$    &   22.6   &     $\text{20.5}_{~\textcolor{cyan}{(-2.1)}}$     \\
(b)  &  47.62      &   $\text{47.37}_{~\textcolor{cyan}{(-0.25)}}$      &   23.4    &  $\text{22.7}_{~\textcolor{cyan}{(-0.7)}}$        \\
(c)  &      49.56    &    $\text{43.09}_{~\textcolor{cyan}{(-6.47)}}$     &     24.3    &   $\text{20.6}_{~\textcolor{cyan}{(-3.7)}}$   \\
(d) &      45.45    &    $\text{42.06}_{~\textcolor{cyan}{(-3.39)}}$     &  21.1     &  $\text{18.9}_{~\textcolor{cyan}{(-2.2)}}$   \\
(e) &      46.34        &    $\text{39.51}_{~\textcolor{cyan}{(-6.83)}}$      &     23.0      &  $\text{17.4}_{~\textcolor{cyan}{(-5.6)}}$        \\ \hline
\end{tabular}
\caption{\textbf{Comparison among various model configurations.} (a) Our default configuration (b) No BigGAN block (c) Self-attention only at bottleneck (8$\times$8 resolution) (d) Two residual blocks (e) Learning rate $2.5e^{-5}$. Samples generated with 250 steps.}
\vspace{-1em}
\label{table:architecture}
\end{table}

\begin{table}[]
\centering
\begin{tabular}{cccc}
\hline
Method                & Schedule    & FID-10k$\downarrow$   & KID-10k$\downarrow$ \\ \hline
\multirow{4}{*}{Base} & 250 uniform & 10.88 & 5.91 \\
                      & 130-60-60   & 10.62  &  5.14    \\
                      & 60-130-60   & 12.23  &  7.54  \\
                      & 500 uniform & 9.74  & 4.48 \\ \hline
Ours                  & 250 uniform & \textbf{8.92}  & \textbf{4.24}  \\ \hline
\end{tabular}
\caption{\textbf{Sweeping sampling schedule.} Schedules expressed as a sequence of integers which are numbers of steps assigned to one-third of the diffusion process. 130-60-60 implies consuming more steps near $t=0$. Modifying the sampling schedule can slightly improve performance, but does not exceed our improvement. 
}

\label{table:sweep}
\end{table}

\section{Related Work}
\label{sec:related}

\subsection{Diffusion-Based Generative Models}
Diffusion models~\cite{sohl2015deep,ho2020denoising,nichol2021improved,dhariwal2021diffusion} and score-based models~\cite{song2019generative,song2020score} are two recent families of generative models that generate data using a learned denoising process. Song et al.~\cite{song2020score} showed that both families can be expressed with stochastic differentiable equations (SDE) with different noise schedules. We note that score-based models may enjoy different hyperparamters of P2 ($\gamma$ and $k$) as noise schedules are correlated with weighting schemes (\cref{eq:vlb}).
Recent studies~\cite{saharia2021image,dhariwal2021diffusion,song2020score} have achieved remarkable improvements in sample quality. However, they rely on heavy architectures, long training and sampling steps~\cite{song2020score}, classifier guidance~\cite{dhariwal2021diffusion}, and a cascade of multiple models~\cite{saharia2021image}. In contrast, we improved the performance by simply redesigning the training objective without requiring heavy computations and additional models. Along with the success in the image domain, diffusion models have also shown effectiveness in speech synthesis~\cite{kong2020diffwave,chen2020wavegrad}.

\subsection{Advantages of Diffusion Models}

Diffusion models have several advantages over other generative models. First, their sample quality is superior to likelihood-based methods such as autoregressive models~\cite{salimans2017pixelcnn++,oord2016conditional}, flow models~\cite{dinh2014nice,dinh2016density}, and variational autoencoders (VAEs)~\cite{kingma2013auto}. Second, because of the stable training, scaling and applying diffusion models to new domains and datasets is much easier than generative adversarial networks (GANs)~\cite{goodfellow2014generative}, which rely on unstable adversarial training. 

Moreover, pre-trained diffusion models are surprisingly easy to apply to downstream image synthesis tasks.
Recent works~\cite{choi2021ilvr,meng2021sdedit} have demonstrated that pre-trained diffusion models can easily adapt to image translation and image editing. Compared to GAN-based methods~\cite{isola2017image,zhu2017unpaired,abdal2019image2stylegan}, they adapt a single diffusion model to various tasks without task-specific training and loss functions. They also show that diffusion models allow stochastic (one-to-many) generation in those tasks, while GAN-based methods suffer deterministic (one-to-one) generations~\cite{isola2017image}.

\subsection{Redesigning Training Objectives}
Recent works \cite{kingma2021variational,song2021maximum,vahdat2021score} introduced new training objectives to achieve state-of-the-art likelihood. However, their objectives suffer from degradation of sample quality and training instability, therefore rely on importance sampling~\cite{song2021maximum,vahdat2021score} or sophisticated parameterization~\cite{kingma2021variational}. Because the likelihood focuses on fine-scale details, their objectives impede understanding global consistency and high-level concepts of images. For this reason, \cite{vahdat2021score} use different weighting schemes for likelihood training and FID training. Our P2 weighting provides a good inductive bias for perceptually rich contents, allowing the model to achieve improved sample quality with stable training.

\section{Conclusion}
\label{sec:conclusion}
We proposed perception prioritized weighting, a new weighting scheme for the training objective of the diffusion models. We investigated how the model learns visual concepts at each noise level during training, and divided diffusion steps into three groups. 
We showed that even the simplest choice (P2) improves diffusion models across datasets, model configurations, and sampling steps. 
Designing a more sophisticated weighting scheme may further improve the performance, which we leave as future work. We believe that our method will open new opportunities to boost the performance of diffusion models. 

\vspace{1em}
\textbf{Acknowledgements:} This work was supported by Institute of Information \& communications Technology Planning \& Evaluation (IITP) grant funded by the Korea government (MSIT) [NO.2021-0-01343, Artificial Intelligence Graduate School Program (Seoul National University)], LG AI Research, Samsung SDS, AIRS Company in Hyundai Motor and Kia through HMC/KIA-SNU AI Consortium Fund, and the BK21 FOUR program of the Education and Research Program for Future ICT Pioneers, Seoul National University in 2022.

{\small 
\bibliographystyle{ieee_fullname}
\bibliography{egbib}
}

\clearpage
\newpage

\twocolumn[
]

\setcounter{section}{0}
\renewcommand\thesection{\Alph{section}}
\setcounter{table}{0}
\renewcommand{\thetable}{\Alph{table}}
\setcounter{figure}{0}
\renewcommand{\thefigure}{\Alph{figure}}
\setcounter{equation}{0}
\renewcommand{\theequation}{\Alph{equation}}

\begin{figure*}[t!]
  \centering
  \includegraphics[width=1.0\linewidth]{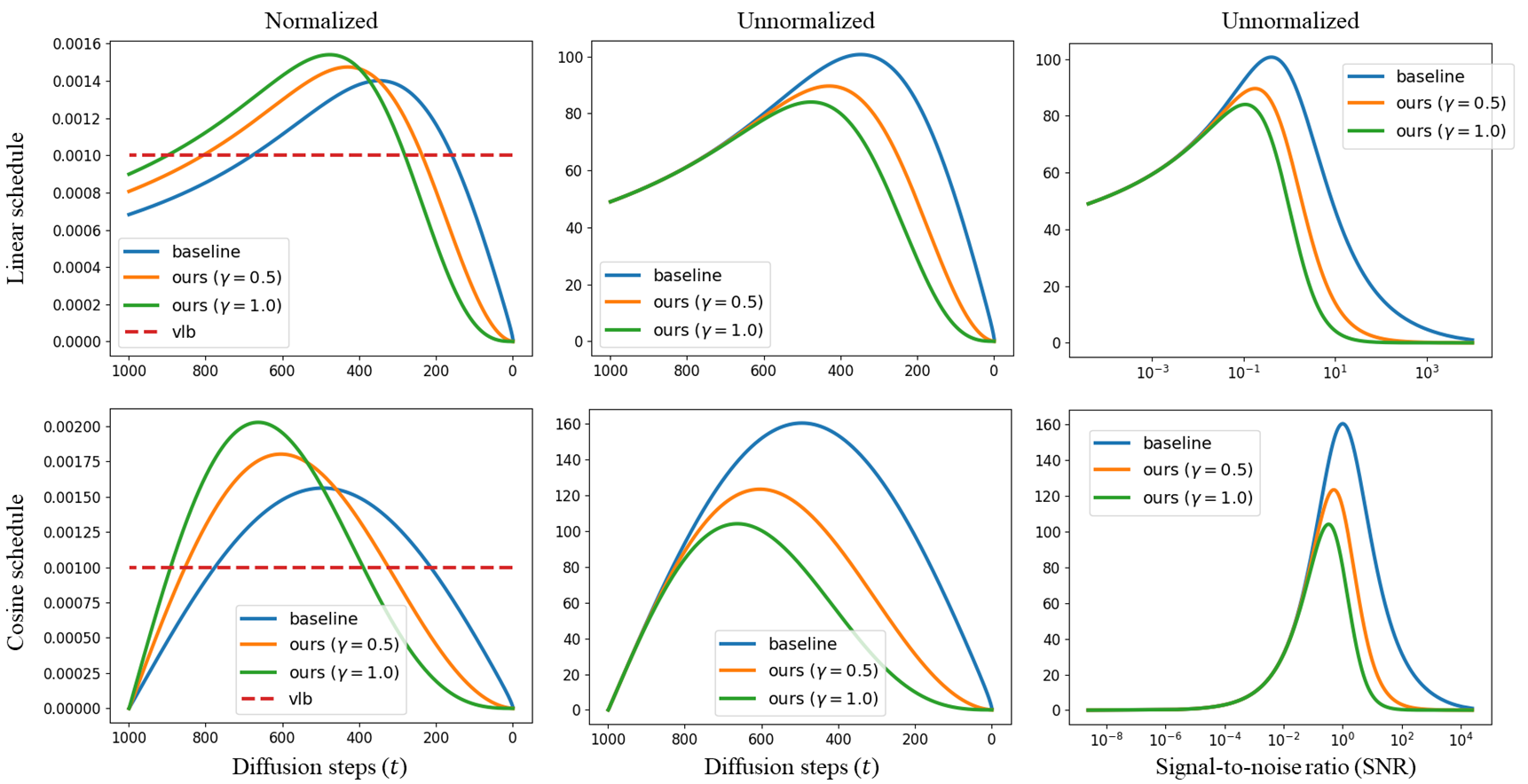}
  \caption{Unnormalized or normalized weights as functions of diffusion steps or signal-to-noise ratio (SNR). Large $t$ and small SNR indicates noisy image $x_t$ near random noise $x_T$, whereas small $t$ and large SNR indicates $x_t$ near a clean image $x_0$.}
  \vspace{6.5em}
  \label{fig:weight}
\end{figure*}

\section{Weighting Schemes}
\subsection{Additional Visualizations}
In the main text, we showed weights of both our new weighting scheme and the baseline, as functions of signal-to-noise ratio (SNR). In \cref{fig:weight} (left), we show weights as functions of time steps ($t$). To exhibit relative changes of weights, we show normalized weights as functions of both time steps (\cref{fig:weight} (middle)) and SNR (\cref{fig:weight} (right)). We normalized so that the sum of weights for all time steps become 1. Normalized weights suggest that larger $\gamma$ suppresses weights at steps near $t=0$ and uplifts weights at larger steps. Note that weights of VLB objective are equal to a constant, as such objective does not impose any inductive bias for training. In contrast, as discussed in the main text, our method encourages the model to learn rich content rather than imperceptible details.

\subsection{Derivations}
In the main text, we wrote the baseline weighting scheme $\lambda_t$ as a funtion of SNR, which characterizes the noise level at each step $t$. Below is the derivation:

\begin{align}\label{eq:derivation}
\lambda _t=~&
(1-\beta _t)(1-\alpha _t)/\beta _t\nonumber \\
=~&(\alpha _t / \alpha_{t-1})(1-\alpha _t)/(1-\alpha _t/\alpha _{t-1})\nonumber \\
=~&\alpha _t(1-\alpha _t)/(\alpha _{t-1}-\alpha _t)\nonumber \\
=~&\alpha _t(1-\alpha _t)/((1-\alpha _t)-(1 - \alpha _{t-1}))\nonumber \\
=~&\frac{\text{SNR}(t)}{(1+\text{SNR}(t))^2}/(\frac{1}{1+\text{SNR}(t)}-\frac{1}{1+\text{SNR}(t-1)})\nonumber \\
=~&\frac{\text{SNR}(t)}{(1+\text{SNR}(t))^2}/\frac{\text{SNR}(t-1)-\text{SNR}(t)}{(1+\text{SNR}(t))(1+\text{SNR}(t-1))}\nonumber \\
=~&\frac{\text{SNR}(t)(1+\text{SNR}(t-1))}{(1+\text{SNR}(t))(\text{SNR}(t-1)-\text{SNR}(t))}\nonumber \\
\approx~&\frac{-\text{SNR}(t)}{\text{SNR}'(t)}~(T\to \infty),
\end{align}
which is a differential of log-SNR($t$) regarding time-step $t$.

\section{Discussions}
\subsection{Limitations}
Despite the promising performances achieved by our method, diffusion models still need multiple sampling steps. Diffusion models require at least 25 feed-forwards with DDIM sampler, which makes it difficult to use diffusion models in real-time applications. Yet, they are faster than autoregressive models which generate a pixel at each step. In addition, we have observed in section 4.3 that our method enables better FID with half the number of steps required by the baseline. Along with our method, optimizing sampling schedules with dynamic programming~\cite{watson2021learning} or distilling DDIM sampling into a single step model~\cite{luhman2021knowledge} might be promising future directions for faster sampling.

\subsection{Broader Impacts}
The proposed method in this work allows high-fidelity image generation with diffusion-based generative models. Improving the performance of generative models can enable multiple creative applications~\cite{choi2021ilvr,meng2021sdedit}. However, such improvements have the potential to be exploited for deception. Works in deepfake detection~\cite{wang2020cnn} or watermarking~\cite{yu2021artificial} can alleviate the problems. Investigating invisible frequency artifacts~\cite{wang2020cnn} in samples of diffusion models might be promising approach to detect fake images.

\section{Implementation Details}

For a given time-step $t$, the input noisy image $x_t$ and output noise prediction $\epsilon$ and variance $\sigma _t$ are images of the same resolution. Therefore, $\epsilon _\theta$ is parameterized with the U-Net~\cite{ronneberger2015u}-style architecture of three input and six output channel dimensions. We inherit the architecture of ADM~\cite{dhariwal2021diffusion}, which is a U-Net with large channel dimension, BigGAN~\cite{brock2018large} residual blocks, multi-resolution attention, and multi-head attention with fixed channels per head. Time-step $t$ is provided to the model by adaptive group normalization (AdaGN), which transforms $t$ embeddings to scales and biases of group normalizations~\cite{wu2018group}. However, for efficiency, we use fewer base channels, fewer residual blocks, and a self-attention at a single resolution (16$\times$16). 

Hyperparameters for training models are in \cref{table:config}. We use $\gamma = 0.5$ for FFHQ and CelebA-HQ as it achieve slightly better FIDs than $\gamma = 1.0$ on those datasets. Models consist of one or two residual blocks per resolution and self-attention blocks at 16$\times$16 resolution or at bottleneck layers of 8$\times$8 resolution. Our default model has only 94M parameters, while recent works rely on large models (larger than 500M)~\cite{dhariwal2021diffusion}. While recent works use 2 or 4 blocks per resolution, we use only one block, which leads to speed-up of training and inference. We use dropout when training on limited data. We trained models using EMA rate of 0.9999, 32-bit precision, and AdamW optimizer~\cite{loshchilov2017decoupled}.

\begin{table*}[]
\centering
\begin{tabular}{lcccccc}
\hline
               & FFHQ, CelebA-HQ         & AFHQ-D                  & CUB, Flowers, MetFaces  & Tab. 3 (b)              & Tab. 3 (c)              & Tab. 3 (d)              \\ \hline
$T$            & 1000                    & 1000                    & 1000                    & 1000                    & 1000                    & 1000                    \\
$\beta_t$      & linear                  & linear                  & linear                  & linear                  & linear                  & linear                  \\
Model Size     & 94                      & 94                      & 94                      & 81                      & 90                      & 132                     \\
Channels       & 128                     & 128                     & 128                     & 128                     & 128                     & 128                     \\
Blocks         & 1                       & 1                       & 1                       & 1                       & 1                       & 2                       \\
Self-attn      & 16, bottle              & 16, bottle              & 16, bottle              & 16, bottle              & bottle                  & 16, bottle              \\
Heads Channels & 64                      & 64                      & 64                      & 64                      & 64                      & 64                      \\
BigGAN Block   & yes                     & yes                     & yes                     & no                      & yes                     & yes                     \\
$\gamma$       & 0.5                     & 1.0                     & 1.0                     & 1.0                     & 1.0                     & 1.0                     \\
Dropout        & 0.0                     & 0.1                     & 0.1                     & 0.1                     & 0.1                     & 0.1                     \\
Learning Rate  & $\text{2e}^{\text{-5}}$ & $\text{2e}^{\text{-5}}$ & $\text{2e}^{\text{-5}}$ & $\text{2e}^{\text{-5}}$ & $\text{2e}^{\text{-5}}$ & $\text{2e}^{\text{-5}}$ \\
Images (M)     & 18, 4.4               & 2.4                     & 4.8, 4.4, 1.6           & 0.8                     & 0.8                     & 0.8                     \\ \hline
\end{tabular}
\caption{Hyperparameters.}
\vspace{6.5em}
\label{table:config}
\end{table*}

\section{Additional Results}
\textbf{Qualitative.  } Additional samples for all datasets mentioned in the paper are in \cref{fig:sample_appendix}.

\begin{figure*}[t!]
  \centering
  \includegraphics[width=1.0\linewidth]{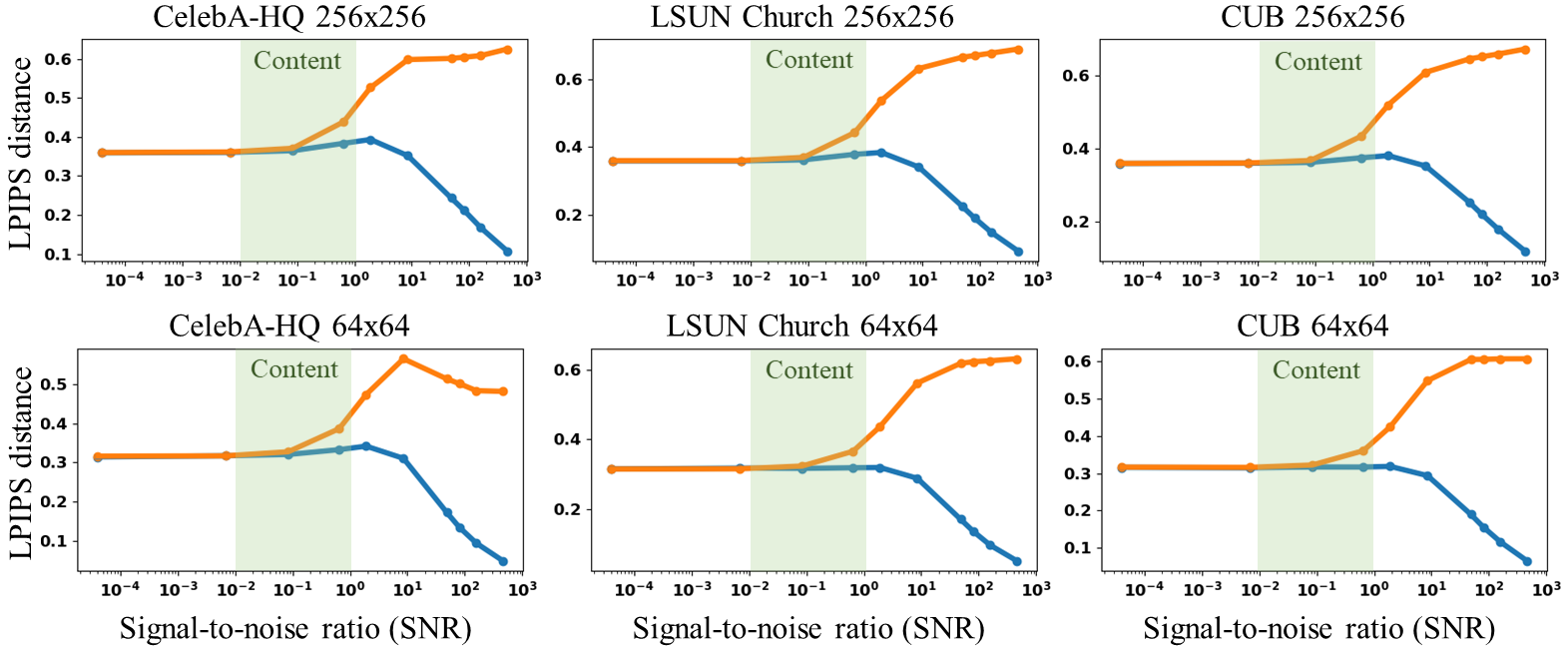}

  \caption{\textbf{Generalization of sec. 3.} Results with CelebA-HQ, LSUN-Church, and CUB at $256^2$ and $64^2$ resolutions.}

  \label{fig:generalize}
\end{figure*}

\begin{figure}[t!]
  \centering
  \includegraphics[width=0.80\linewidth]{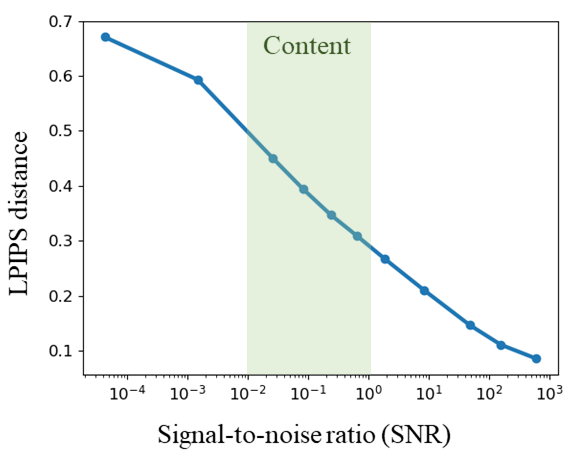}
  \caption{\textbf{Stochastic reconstruction.} Perceptual distance between input and reconstructed image as a function of signal-to-noise ratio, measured with random 200 images from FFHQ.}
  \label{fig:recon_appendix}
\end{figure}

\begin{figure*}[t!]
  \centering
  \includegraphics[width=0.90\linewidth]{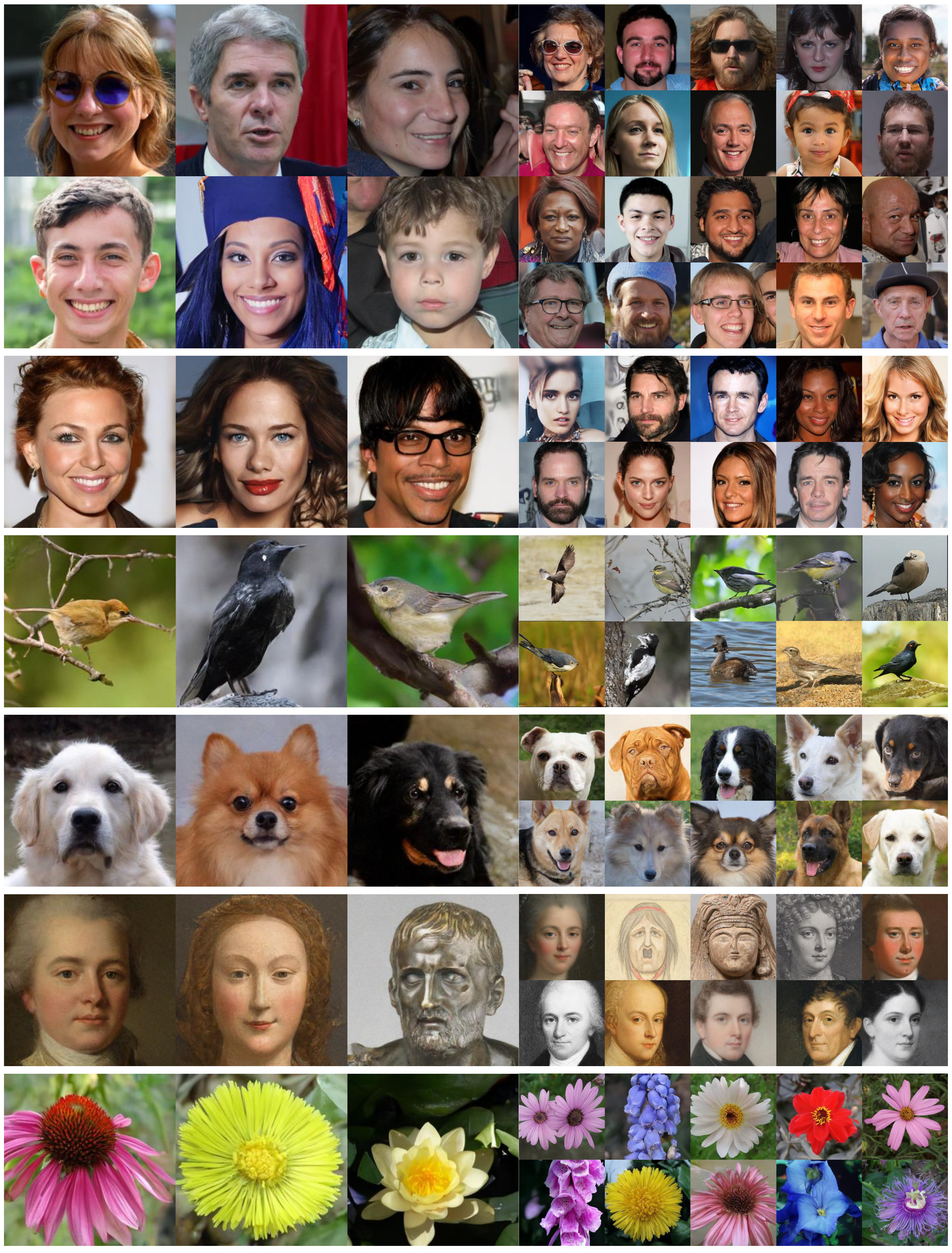}
  \caption{Additional samples generated with our models traind on various datasets.}
  \label{fig:sample_appendix}
\end{figure*}

\textbf{Quantitative.  } In Fig. 1 of the main text, we measured perceptual distances to investigate how the diffusion process corrupts perceptual contents. In Fig. 2, we qualitatively explored what a trained model learned at each step (Fig. 2). Here, we reproduce Fig. 1 at various datasets and resolutions in \cref{fig:generalize} and show the quantitative result of Fig. 2 in \cref{fig:recon_appendix}. These results indicate that our investigation in Sec. 3.1 holds for various datasets and resolutions.

\newpage

\end{document}